\documentclass[runningheads]{llncs}

\usepackage[T1]{fontenc}
\usepackage{nicefrac}
\usepackage{subcaption}
\usepackage{booktabs}
\usepackage{float}
\usepackage{graphicx}
\usepackage{algorithm}
\usepackage{algorithmic}

\usepackage{amsmath}

\usepackage{booktabs, multirow, colortbl, caption}
\usepackage[table,xcdraw]{xcolor}

\begin{document}

\title{Understanding trade-offs in classifier bias with quality-diversity optimization: An application to talent management}  

\titlerunning{Understanding trade-offs in classifier bias with quality-diversity optimization}

\author{Catalina M. Jaramillo\inst{1}\orcidID{0000-0002-2867-3405} \and
Paul Squires\inst{1}\orcidID{0009-0008-5209-6776} \and
Julian Togelius\inst{1}\orcidID{0000-0003-3128-4598}}

\authorrunning{C.M. Jaramillo et al.}
%
\institute{New York University, New York, NY, USA}

\maketitle

\begin{abstract}
  Fairness, the impartial treatment towards individuals or groups regardless of their inherent or acquired characteristics~\cite{mehrabi2021survey}, is a critical challenge for the successful implementation of Artificial Intelligence (AI) in multiple fields like finances, human capital, and housing. A major struggle for the development of fair AI models lies in the bias implicit in the data available to train such models. Filtering or sampling the dataset before training can help ameliorate model bias but can also reduce model performance and the bias impact can be opaque. In this paper, we propose a method for visualizing the biases inherent in a dataset and understanding the potential trade-offs between fairness and accuracy. Our method builds on quality-diversity optimization, in particular Covariance Matrix Adaptation Multi-dimensional Archive of Phenotypic Elites (MAP-Elites). Our method provides a visual representation of bias in models, allows users to identify models within a minimal threshold of fairness, and determines the trade-off between fairness and accuracy.

\end{abstract}

\section{Introduction}

Bias exists in various forms throughout society and can be learned by AI based automated decision-making systems, which learn to recognize patterns and consequently replicate biases ingrained in those patterns~\cite{hajian2016algorithmic}. While not inherently negative, biases in AI models can exacerbate negative impacts on individuals and society, eroding trust in AI ~\cite{mehrabi2021survey,huang2022overview,pessach2022review}. 
Numerous examples highlight how biased AI technologies in areas like hiring, healthcare, and criminal justice have caused harm and raised concerns about maintaining societal biases~\cite{liem2018psychology,schwartz2022towards}. Research has found how discrimination based on gender, race, age, and other characteristics can be perpetuated by the use of biased AI models~\cite{sweeney2013discrimination,chen2023comprehensive,hoffman2019artificial}. 

Any dataset allows for the learning of an almost arbitrarily large number of models with similar accuracy. This situation has been defined as multiplicity. Meyer\cite{meyer2023dataset}, describes dataset multiplicity as the multiplicity introduced in the model by data inaccuracies (like errors in data collection, data being nonrepresentative of the population, bias, etc). Black et al.\cite{black2022model} shows how models with similar performance can have different results from the fairness perspective, and how such multiplicity can be caused by factors like feature selection, training sampling, model class, parameters initialization. Watson et al.\cite{watson2023multi} add the target output selection as another possible cause.

Algorithm and dataset biases combine in complex ways to create biased learned models. When applying ML to real world problems, bias can manifest in undesirable ways~\cite{mehrabi2021survey}. The debate about the societal effects of biased models is usually oriented towards model bias induced by datasets~\cite{hellstrom2020bias}. Technically, bias can also come from e.g. algorithmic details or architectural choices. For example, algorithms that predict job success and are biased towards a particular gender or ethnicity are clearly problematic. For example, if an algorithm that is used to screen job applicants or provide decision support to a hiring manager is trained on hiring decisions that were made in a biased way (such as favoring men over women), its screening or recommendations will likely incorporate those biases. It becomes extremely important to find ways to not only identify such biases, but also selectively counteract them while minimizing the impact on model performance, as measured by accuracy \cite{liem2018psychology}. 

A method can be developed to find sets of trained models with good accuracy, but that are different according to their expression of specific biases. Such method could be used to identify models that perform well while minimizing these biases. Such a set of models could also help us understand the underlying trade-offs, if any, between model performance and biases. If the underlying dataset is biased, the best model of data is very likely also biased; we may instead want the best model that is unbiased according to some measure, or that has no more than a certain amount of bias. We could call this the bias-performance trade-off. Visualizing and understanding this trade-off, however, requires that the set of models is identified so that it becomes easy to find the relevant models. 

In this paper, we introduce a new method for understanding classifier biases induced by data, and helping to select a model that has the right trade-off between performance and bias. Our method is based on quality diversity (QD) search, a relatively recent family of algorithms that finds a large number of solutions that vary in systematic ways. In other words, we embrace model multiplicity, and provide a way of helping select between models.

\section{Background}

Our method builds on quality-diversity search and is situated within a set of approaches to address bias in machine-learned models; in this paper, it is applied to a talent management problem. We give a brief background on each topic.

\subsection {Bias in talent management and the 80\% rule}

Our main example here is taken from talent management, specifically promotion, using a a dataset of promotion decisions. Here, we provide some background on bias in promotion and the need for visualizing balance and finding sufficiently fair yet performant models.

The issue of bias and fairness toward workers has been of keen interest to employers for many decades. In the US, a greater urgency and attention to bias and fairness was ushered in by the passage of the Civil Rights laws of the 1960s and 1970s and subsequent extensions and revisions. In the years following the passage of these laws there has been a great deal of litigation at all levels of the judicial system, including many US Supreme Court cases (Griggs vs Duke Power, Albemarle vs Moody, Watson vs Fort Worth Bank, Connecticut vs Teal)~\cite{griggs,albemarle,watson,connecticut}. The litigation clarified the law and established “rules” for assessing bias and fairness. One such rule is the 80\% or \nicefrac{4}{5} rule.  According to the EEOC’s Uniform Guidelines on Employee Selection Procedures (1978) the 80\% rule is a criterion to identify adverse impact and is defined as, “A selection rate for any race, sex, or ethnic group which is less than four-fifths (\nicefrac{4}{5}) (or eighty percent) of the rate for the group with the highest rate will generally be regarded by the Federal enforcement agencies as evidence of adverse impact, while a greater than four-fifths rate will generally not be regarded by Federal enforcement agencies as evidence of adverse impact.”

Employers then are challenged with deploying employee selection and promotion procedures and cutoff scores (tests, GPAs, interview ratings, etc) in a way that accurately identifies talented and deserving individuals while at the same time avoids adverse impact, that is, failing to meet the 80\% rule. The problem may be framed as optimizing the balance between fairness and hiring and rewarding top performers. So for example if an employer wishes to set a cut-off of 75\% on an employment test in which higher scores predict better job performance and the pass rate for females is less than 80\% of the pass rate for male, what should the employer do? The complexity of the problem becomes apparent when considering cutoff scores for multiple screening procedures and their potential adverse impact for multiple groups (gender, ethnicity, age, etc.)~\cite{CFR}.

\subsection{Methods for handling bias in machine learning}

Research has explored different possible sources of bias and its negative impact in the application of the models. Suresh et al.~\cite{suresh2021framework} show how harm can be introduced in different steps of the ML cycle, going from data collection and preparation, model development and evaluation to results deployment.

Multiple approaches have been proposed to mitigate unfairness. Chen et al.~\cite{chen2023comprehensive} evaluate 17 bias mitigation methods, including Optimized Pre-processing, Learning Fair Representation, Disparate Impact Remover, Adversarial Debiasing, Meta Fair Classifier, and Reject Option Classification, among others. Their findings show that bias fairness is improved in about 46\% of the studied cases, but there is also a significant negative impact in model accuracy in 66\% of the scenarios, and actually in 25\% of scenarios fairness is negatively affected by the mitigation method.
Zhang et al.~\cite{zhang2021fairer,zhang2022mitigating,zhang2024fairness} proposed a multi-objective evolutionary method where learning uses objectives that combine accuracy and fairness metrics; later, they included ensemble learning and a dynamic adaptative selection of used fairness metrics.

Another important factor to consider is the difficulty to understand and measure the trade-off
between accuracy and fairness~\cite{speicher2018unified} and the complexity of different approaches suggested to manage its impact on models~\cite{li2024triangular}. By finding a set of models with diverse results from the bias perspective, instead of a single model optimization approach as proposed by Zhang et al. ~\cite{zhang2021fairer,zhang2022mitigating,zhang2024fairness}, our method allows for a better understanding of the mentioned trade-off and the selection of the model that better fits the given restrictions.

\subsection{Quality-diversity search}

The quality-diverse (QD) framework is a somewhat recent family of stochastic search/optimization algorithms, where optimization in favor of an objective or fitness function is considered to be only one part of the story~\cite{pugh2016quality,cully2017quality}. QD algorithms instead find diverse sets of solutions, according to some diversity measure or set of diversity measures. In particular, we will focus our investigation on a version of the Map-Elites algorithm, which is a stochastic illumination algorithm that finds a set of solutions to cover a map defined by one or more descriptors~\cite{mouret2015illuminating}. 
Functionally, Map-Elites can be thought of as an evolutionary algorithm where the population is spread out on a map according to what descriptor values each individual has. The map is divided into cells, where each cell contains the best solution found so far within a particular descriptor interval. This makes it easy to find good solutions that exhibit desired behaviors. For example, if searching for robot gaits, Map-Elites can generate a map of gaits where the descriptors correspond to which legs are used, and a good robot gait that fits a specific leg configuration can easily be found~\cite{cully2015robots}.


Map-Elites, like other forms of evolutionary computation, is a stochastic search/optimization algorithm~\cite{cully2017quality}. This means that it is at efficiency disadvantage compared to methods for model optimization tailored to specific learner representations, such as gradient descent (for neural network training) or information gain maximization (for decision tree learning). However, it has the advantage that stochastic search can be applied to optimize anything that has parameters and an objective function. Common learner representations such as neural networks and decision trees can be evolved with good results \cite{wang2011novel}, though this process is often less computationally efficient than using model-specific learning methods.

The particular version of Map-Elites we use here is Covariance Matrix Adaptation MAP-Elites (CMA-ME), which allows Map-Elites to search continuous spaces more efficiently as shown in Algorithm~\ref{alg:cma-me}~\cite{fontaine2020covariance}. CMA-ME adapts the covariance matrix adaptation method from Covariance Matrix Adaptation Evolution Strategies to a QD setting~\cite{hansen2001completely}.

\begin{algorithm}[h]
\caption{Covariance Matrix Adaptation MAP-Elites}
\label{alg:cma-me}
\begin{algorithmic}[1]
\STATE \textbf{Input:} An evaluation function $evaluate$ which computes a behavior characterization and fitness, and a desired number of solutions $n$.
\STATE \textbf{Result:} Generate $n$ solutions storing elites in a map $M$.
\STATE Initialize population of emitters $E$.
\FOR{$i = 1$ to $n$}
    \STATE Select emitter $e$ from $E$ which has generated the least solutions out of all emitters in $E$.
    \STATE $x_i \gets \text{generate\_solution}(e)$
    \STATE $(\beta_i, \text{fitness}) \gets evaluate(x_i)$
    \STATE $\text{return\_solution}(e, x_i, \beta_i, \text{fitness})$
\ENDFOR

\end{algorithmic}
\end{algorithm}

\section{Methodology}

In this paper, we propose the use of QD methods, specifically Map-Elites, to find sets of models that vary systematically in selected bias dimensions. We operationalize biases as the overrepresentation of certain features among the predicted instances of a model, and use degrees of biases as descriptors for CMA-ME. 

Additionally, we use a fairness definition based on the predicted outcomes of the classifier, focusing on group fairness, which requires that all groups within a protected attribute have a similar probability of being assigned to the positive predicted class~\cite{verma2018fairness}.

The first step is to obtain a baseline model: a simple multi-layer neural network is created, using hyper-parameter optimization and 10-fold stratified cross-validation, to identify the best model. The resulting configuration (number of layers and number of nodes) is implemented in the neural network used in the experiments.

Next, the weights of a set of neural networks with the defined architecture are evolved, and their fitness values are calculated as the accuracy of the network forward pass prediction against the real outcome for the whole dataset. 

The above-mentioned accuracy is used as the output value to be maximized, while the ratio of positive classification between the two groups present in each protected attribute are used as descriptors of the model (for example the ratio of female positive prediction rate vs male positive prediction rate for gender attribute, or the ratio of young positive rate vs old positive rate for age.) See Algorithm \ref{alg:fitness}.

\begin{algorithm}[H]
\caption{Fitness Function}
\label{alg:fitness}
\begin{algorithmic}[1]
\STATE \textbf{Input:} \texttt{weights}, \texttt{nn\_architecture}, \texttt{cases}, \texttt{cases\_xa}, \texttt{cases\_xb}, \texttt{cases\_ya}, \texttt{cases\_yb}, \texttt{labels}.
\STATE \textbf{Output:} \texttt{accuracy}, \texttt{ratio\_x}, \texttt{ratio\_y}.

\STATE \textbf{Step 1: Compute Overall Accuracy}
\STATE Perform a forward pass on \texttt{cases} with \texttt{weights} and \texttt{nn\_architecture}.
\STATE Compute \texttt{accuracy}:
\[
\texttt{accuracy} \gets \frac{\texttt{correct predictions}}{\texttt{total cases}}
\]

\STATE \textbf{Step 2: Compute Group Ratios}
\STATE For each group (\texttt{cases\_xa}, \texttt{cases\_xb}, \texttt{cases\_ya}, \texttt{cases\_yb}), compute mean predictions:
\[
\texttt{mean\_xa}, \texttt{mean\_xb}, \texttt{mean\_ya}, \texttt{mean\_yb}
\]
\STATE Compute fairness ratios:
\[
\texttt{ratio\_x} \gets \frac{\texttt{mean\_xa}}{\texttt{mean\_xb} + \epsilon}, \quad
\texttt{ratio\_y} \gets \frac{\texttt{mean\_ya}}{\texttt{mean\_yb} + \epsilon}
\]

\STATE \textbf{Step 3: Return Fitness Results}
\STATE \textbf{Output:} (\texttt{accuracy}, (\texttt{ratio\_x}, \texttt{ratio\_y}))
\\
\end{algorithmic}
\end{algorithm}

A two-dimensional map was built, showing the best accuracy reached for each cell. The cells in the map are delimited by the ratios of two protected attributes present in the dataset.

\begin{figure}[h]
    \centering
    \includegraphics[width=.7\textwidth]{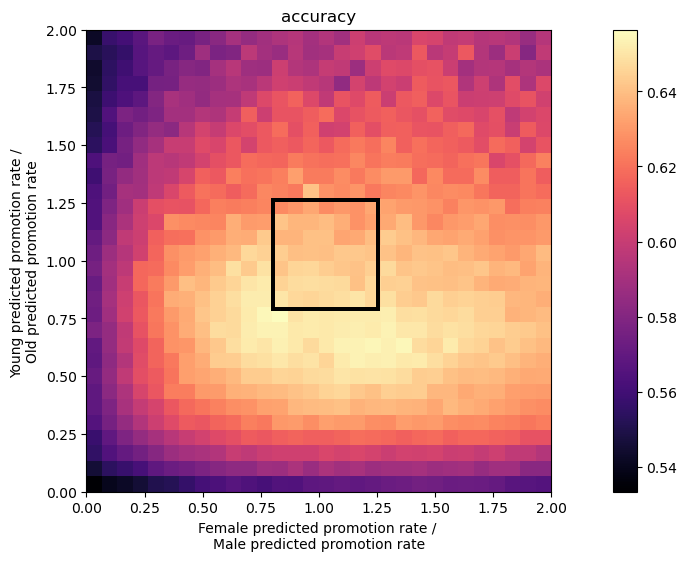}\hfill
    \caption{Fair zone within the map}
    \label{fairzone}
\end{figure}
    
From the map, we identified the model with best accuracy, and its protected attributes' positive ratios; along with the best accuracy for a model within the "fair zone", and the respective positive ratios. The "fair zone" is  defined by the area between 0.8 and 1.25 in the positive ratios based on the 80\% rule described before, as shown in Figure \ref{fairzone}. A model that is located closer to (1,1) (where positive prediction is equally likely along both groups within each demographic dimension) has a lower learned bias. The bigger the distance from (1,1), the higher the bias. We used Euclidean distance between (1,1) and the coordinates given by the model descriptors to compare the bias of different models.

Two datasets were used in this research. In the first one we apply the method to predicting promotions in a company using the Promotion Dataset~\cite{dataset1}, a task with high relevance for ML applications in human resources. In the second case, we used the Adult Dataset~\cite{adult_2}, a dataset well known for fairness research.

For each case we used two protected attributes in the dataset and selected the best model with an acceptable degree of bias, given by the the 80\% rule. By comparing the selected model with the most accurate one in the whole map, the method allows to visualize the trade-off between bias and model performance. The map also reflects how the bias in the data impacts the results, by simply displaying the array of most accurate models and color-coding each cell for accuracy.

\begin{table}[ht]
    \caption{Methods}
    \label{methods}
    \centering
    \begin{tabular}{lcc}
        \toprule
        & \textbf{Promotion Dataset} & \textbf{Adult Dataset} \\
        \midrule
        N    & 54808 & 48842 \\
        Protected attributes    & Gender, Age & Gender, Race \\
        Number features & 14 & 55 \\
        Baseline Accuracy   & 0.6628 & 0.8416 \\
        \midrule
        \textbf{NN Architecture} & \textbf{} & \textbf{} \\
        \midrule
        NN Hiden layers & 2 & 2 \\
        NN Nodes  & [35,15] & [64,32] \\
        NN Activation function & Leaky  Relu & Leaky  Relu \\
        \bottomrule
    \end{tabular}
\end{table}

The used code relies on the pyribs~\cite{pyribs} library. Details about the CMA-ME parameters and code can be found at \url{https://github.com/catajara/Classif-Bias-Tradeoffs-QD}

\subsection{Data}
The Promotion Dataset, is an open source dataset consisting of individuals that were either promoted or not promoted and a set of features including department in the organization, region, education, gender, training, age, performance, and tenure. Categorical variables were encoded using binary dummy variables. We used sampling to generate different scenarios from the data, to evaluate how the presence or absence of bias in the data impacts the results. Random sampling was used to obtain 4 datasets with a defined bias. In all cases the total percentage of promoted individuals is set close to 50\%. Given the age distribution in the data, a 34 years old threshold was used to define the young vs. old groups.
Rows 1 to 4 in Table \ref{dataset_info_table} show the data used, including the number of cases for each group and Row 1 to 4 in Table \ref{positive_rate_table} show the promotion rate for the group:

\begin{enumerate}

\item Promotion Unbiased dataset, with 50\% promotion rate for all the categories.
\item Promotion dataset with same proportion for male and female, and for young and old; with higher promotion rate for male cases in both age groups. 
\item Promotion dataset with higher sample size for male cases, and similar proportion for young and old groups; with higher promotion rate for male in both age groups.
\item Promotion dataset with same proportion for male and female cases, and similar proportion for young and old; with higher promotion rate for male and young cases (67\% compared with 33\% for male and old), and higher promotion rate for female and old (67\%) vs. female and young cases (33\%).
\end{enumerate}

Adult Dataset, predicts annual income exceeding of \$50K/yr using census data. Features include age, gender, race, education,  occupation, among others. This dataset has bias for both protected attributes (gender and race in this case). Random sampling was used to obtain a stratified sample with similar distribution to the original dataset and a unbiased sample.
Rows 5 and 6 in Tables \ref{dataset_info_table} and \ref{positive_rate_table} show the sample details for Adult dataset:

\begin{enumerate}
    \setcounter{enumi}{4}
\item Adult unbiased dataset , with 50\% promotion rate for all the categories.
\item Adult stratified sample, with higher number cases for male and for white, and higher positive rate for male and for white.
\end{enumerate}

\begin{table}[H]
    \centering
    \caption{Dataset Information}
    \label{dataset_info_table}
    \resizebox{1\textwidth}{!}{%
    \begin{tabular}{c|l|ccccc}
        \toprule
        \textbf{No.} & \textbf{Dataset} & \textbf{Total (n)} & \textbf{Male (n)} 
        & \textbf{Female (n)} & \textbf{Young/} & \textbf{Old} \\[-1ex]
        & & & & & \textbf{White (n)} & \textbf{Other (n)} \\[-1ex]
        \midrule
        1 & Unbiased & 5728 & 2864 & 2864 & 2864 & 2864 \\
        2 & Male Biased & 4384 & 2192 & 2192 & 2192 & 2192 \\
        3 & Higher Male Sample & 6576 & 4384 & 2192 & 3288 & 3288 \\
        4 & Cross Biased & 4508 & 2254 & 2254 & 2254 & 2254 \\
        \hline 
        5 & Adult Unbiased & 1680 & 840 & 840 & 840 & 840 \\
        6 & Adult Stratified & 13565 & 9224 & 4341 & 11666 & 1899 \\
        \bottomrule
    \end{tabular}
    }
\end{table}

\begin{table}[H]
    \centering
    \caption{Positive Rate by Subgroup}
    \label{positive_rate_table}
    \resizebox{0.9\textwidth}{!}{%
    \begin{tabular}{c|l|ccccc}
        \toprule
        \textbf{No.} & \textbf{Dataset} & \textbf{All} & \textbf{Male} 
        & \textbf{Female} & \textbf{Young/White} & \textbf{Old/Other} \\
        \midrule
        1 & Unbiased & 0.50 & 0.50 & 0.50 & 0.50 & 0.50 \\
        2 & Male Biased & 0.50 & 0.65 & 0.35 & 0.50 & 0.50 \\
        3 & Higher Male Sample & 0.56 & 0.67 & 0.35 & 0.56 & 0.56 \\
        4 & Cross Biased & 0.50 & 0.50 & 0.50 & 0.50 & 0.50 \\
        \hline 
        5 & Adult Unbiased & 0.50 & 0.50 & 0.50 & 0.50 & 0.50 \\
        6 & Adult Stratified & 0.25 & 0.31 & 0.11 & 0.26 & 0.16 \\
        \bottomrule
    \end{tabular}
    }
\end{table}

\section{Results}

For each experiment, we recorded several metrics. First, we generated a heat map displaying the best accuracy achieved for attributes' bias descriptors combination. The heat map was organized into 30 bins ranging from 0 to 2.
Additionally, we compiled a table containing various metrics:
\begin{itemize}
    \item Accuracy
    \item Ratio of predicted positive rates for both groups withing each protected attribute
    \item Deviation -Euclidian distance from (1,1)- for the model with the highest accuracy among all models in the map (best model)
    \item Deviation for the model with the highest accuracy located within the fair zone, which ranged from 0.8 to 1.25 for both descriptors (best fair).
\end{itemize}

\begin{figure}[h]
  \centering
  \begin{minipage}[c]{0.45\textwidth} 
    \centering
    \includegraphics[width=0.9\textwidth]{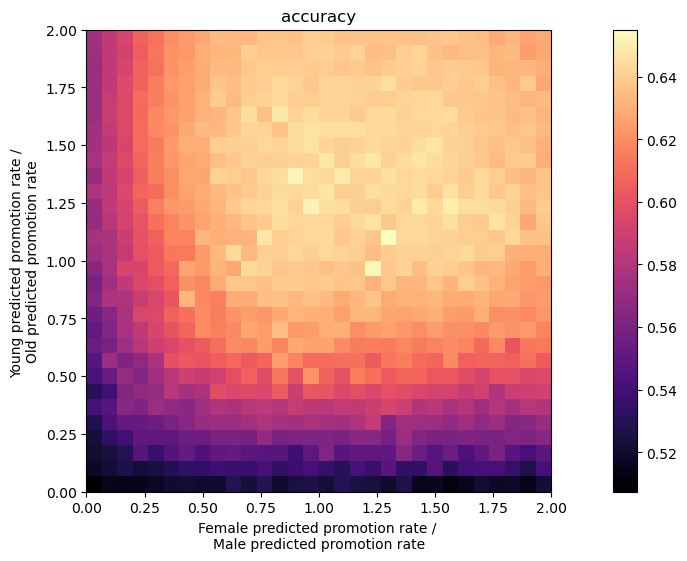} 
    \captionof{figure}{Unbiased sample map}
    \label{unbiasfig}
  \end{minipage}
  \hspace{0.05\textwidth} 
  \begin{minipage}[c]{0.45\textwidth} 
    \centering
    \captionof{table}{Unbiased sample results}
    \label{unbiased results}
    \begin{tabular}{lll}
      \toprule
      & Best model & Best fair \\
      \midrule
      Accuracy & 0.6550 & 0.6444 \\
      Fem/Male & 1.2695 & 0.8365 \\
      Young/Old & 1.0994 & 1.1553 \\
      \midrule
      Deviation & 0.2872 & 0.2255 \\
      \bottomrule
    \end{tabular}
  \end{minipage}
\end{figure}

Following we present the metrics collected for each experiment:

\subsection {Unbiased sample (Figure \ref{unbiasfig}, Table \ref{unbiased results}).}

In this sample (see Row 1 Table \ref{positive_rate_table}), we have the same number of promoted individuals in all the groups: male, female, old and young individuals, making the promotion rate 50\% for all cases.

The accuracy is 0.0106 lower for the best model in the fair zone vs. the best model in the map, while the distance from (1,1) is reduced from 0.2872 to 0.2255.

\begin{figure}[h]
  \centering
  \begin{minipage}[c]{0.45\textwidth} 
    \centering
    \includegraphics[width=0.9\textwidth]{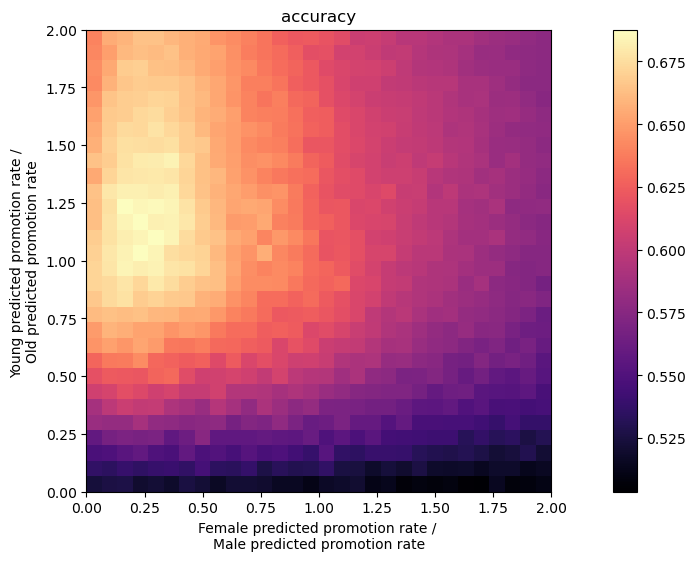} 
    \caption{Male biased sample map}
    \label{malebiasfig}
  \end{minipage}
  \hspace{0.05\textwidth} 
  \begin{minipage}[c]{0.45\textwidth} 
    \centering
    \captionof{table}{Male biased sample results}
    \label{male biased results}
    \begin{tabular}{lll}
      \toprule
      & Best model & Best fair \\
      \midrule
      Accuracy & 0.6877 & 0.6476 \\
      Fem/Male & 0.2400 & 0.8344 \\
      Young/Old & 1.1573 & 1.0990 \\
      \midrule
      Deviation & 0.7761 & 0.1929 \\
      \bottomrule
    \end{tabular}
  \end{minipage}
\end{figure}

\subsection {Male biased sample (Figure \ref{malebiasfig}, Table \ref{male biased results})}

While the number of male and female, and the number of young and old individuals is equally split, the proportion of males in the sample that are promoted is 65\% and 35\% for females. Regarding age, both groups have a similar promotion rate of 50\% (Row 2 Table \ref{positive_rate_table}.)\\

The best model is 0.0401 points more accurate than the best model in the fair zone. 
We observe a higher distance from (1,1) for the most accurate model (0.7761) and that in that model the promotion rate for women is the 24\% of the promotion rate for men. By using the best fair model, distance from (1,1) is reduced to 0.1929 and the ratio between women and men promotion rate is increased to 0.8344.

\begin{figure}[h]
  \centering
  \begin{minipage}[c]{0.45\textwidth} 
    \centering
    \includegraphics[width=0.9\textwidth]{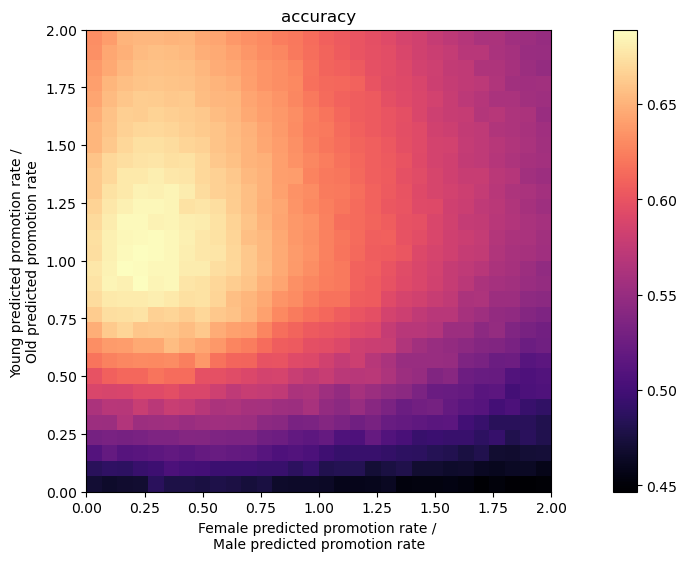} 
    \caption{Male biased and oversampled sample map}
    \label{malebiasoverfig}
  \end{minipage}
  \hspace{0.05\textwidth} 
  \begin{minipage}[c]{0.45\textwidth} 
    \centering
    \captionof{table}{Male biased and oversampled sample results}
    \label{male biased over results}
    \begin{tabular}{lll}
      \toprule
      & Best model & Best fair \\
      \midrule
      Accuracy & 0.6892 & 0.6461 \\
      Fem/Male & 0.2575 & 0.8202 \\
      Young/Old & 0.8808 & 1.0491 \\
      \midrule
      Deviation & 0.7520 & 0.1863 \\
      \bottomrule
    \end{tabular}
  \end{minipage}
\end{figure}

\subsection {Male biased sample, male oversampling (Figure \ref{malebiasoverfig}, Table \ref{male biased over results})}

For this case, we wanted to observe the effect of having a higher promotion rate for males and a higher number of male cases represented in the dataset (Row 2 Tables \ref{dataset_info_table} and  \ref{positive_rate_table}), a condition found in many real life situations. Both age groups have a similar promotion rate.

The distribution of models is similar to the distribution obtained in the second experiment.
In this case, the trade-off in accuracy between the best model and the best fair one is 0.0431, and the ratio for female vs male promotion rate is 0.2575. The distance from (1,1) is 0.7220. This distance is reduced to 0.1863 when we use the best fair model and the female vs male promotion ratio is increased to 0.8202. 

\begin{figure}[h]
  \centering
  \begin{minipage}[c]{0.45\textwidth} 
    \centering
    \includegraphics[width=0.9\textwidth]{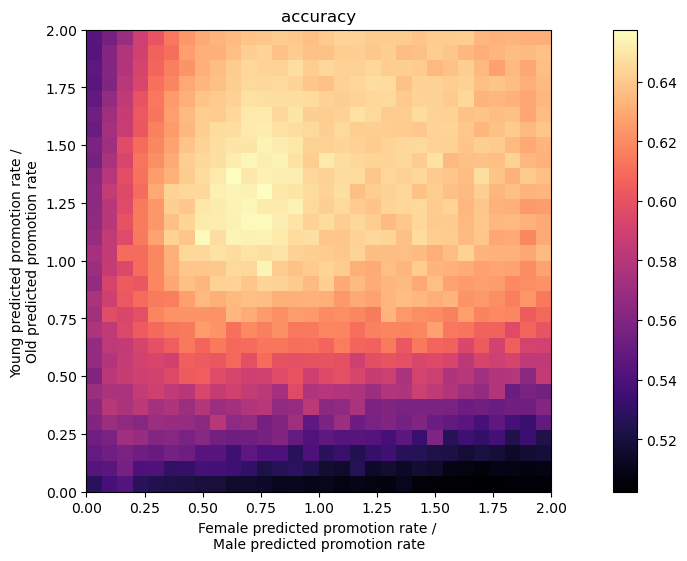} 
    \caption{Cross biased sample map}
    \label{crossbiasfig}
  \end{minipage}
  \hspace{0.05\textwidth} 
  \begin{minipage}[c]{0.45\textwidth} 
    \centering
    \captionof{table}{Cross biased sample results}
    \label{cross biased results}
    \begin{tabular}{lll}
      \toprule
      & Best model & Best fair \\
      \midrule
      Accuracy & 0.6657 & 0.6586 \\
      Fem/Male & 0.8173 & 0.9313 \\
      Young/Old & 1.3600 & 1.0616 \\
      \midrule
      Deviation & 0.4037 & 0.0922 \\
      \bottomrule
    \end{tabular}
  \end{minipage}
\end{figure}

\subsection {Cross biased sample, male young and female old (Figure \ref{crossbiasfig},Table \ref{cross biased results}).}
In this experiment we used a dataset that doesn't have bias considering each dimension independently (male vs. female or young vs old). But, when we combined the dimensions, there is bias: individuals that are male and old have a higher promotion rate, so do individuals that are female and young (Row 4 Table \ref{dataset_info_table})

The accuracy difference between the best model and the best fair model is 0.0071. And the distance is reduced from 0.4037 to 0.0922.

\begin{figure}[h]
  \centering
  \begin{minipage}[c]{0.45\textwidth} 
    \centering
    \includegraphics[width=1.1\textwidth]{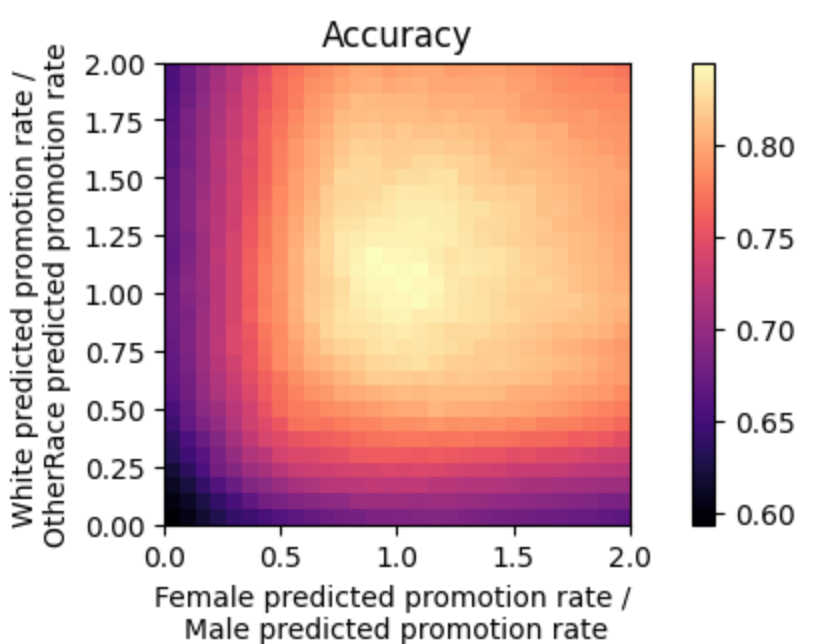} 
    \caption{Adult unbiased sample map}
    \label{adultunbiased}
  \end{minipage}
  \hspace{0.05\textwidth} 
  \begin{minipage}[c]{0.45\textwidth} 
    \centering
    \captionof{table}{Adult unbiased sample results}
    \label{adult unbiased results}
    \begin{tabular}{lll}
      \toprule
      & Best model & Best fair \\
      \midrule
      Accuracy & 0.8452 & 0.8452 \\
      Fem/Male & 0.9341 & 0.9341 \\
      White/Other & 1.0853 & 1.0853 \\
      \midrule
      Deviation & 0.1078 & 0.1078 \\
      \bottomrule
    \end{tabular}
  \end{minipage}
\end{figure}

\subsection {Adult dataset, unbiased sample (Figure \ref{adultunbiased},Table \ref{adult unbiased results}).}
In this experiment we used a unbiased sample from the Adult dataset (Row 5 Table \ref{positive_rate_table})

The best model in the map is within the "fair zone".

\begin{figure}[h]
  \centering
  \begin{minipage}[c]{0.45\textwidth} 
    \centering
    \includegraphics[width=1.1\textwidth]{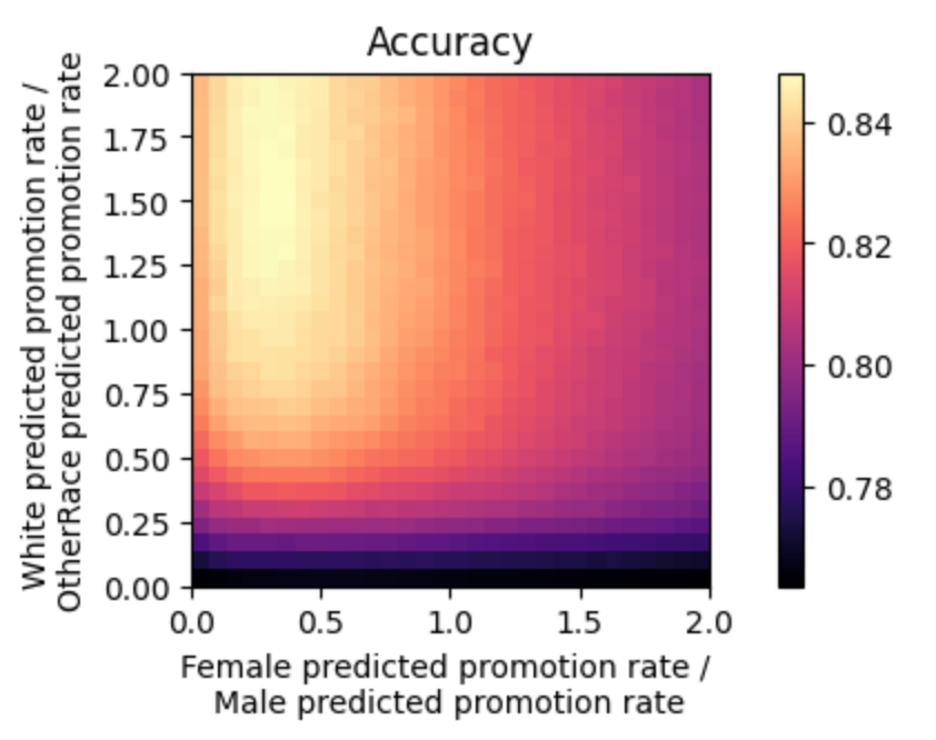} 
    \caption{Adult stratified sample map}
    \label{aacrossbiasfig}
  \end{minipage}
  \hspace{0.05\textwidth} 
  \begin{minipage}[c]{0.45\textwidth} 
    \centering
    \captionof{table}{Adult stratified sample results}
    \label{adult stratified results}
    \begin{tabular}{lll}
      \toprule
      & Best model & Best fair \\
      \midrule
      Accuracy & 0.8481 & 0.8331 \\
      Fem/Male & 0.3443 & 0.8014 \\
      White/Other & 1.5121 & 1.1930 \\
      \midrule
      Deviation & 0.8320 & 0.2769 \\
      \bottomrule
    \end{tabular}
  \end{minipage}
\end{figure}

\subsection {Adult dataset, stratified sample (Figure \ref{aacrossbiasfig},Table \ref{adult stratified results}).}
In this experiment we used a stratified sample from the Adult dataset that corresponds to the distribution of positive outcomes for both groups in the protected attributes (gender and race). See Row 6 Tables \ref{dataset_info_table} and \ref{positive_rate_table}).

The most accurate models are distributed in a similar way than the bias in the dataset. The accuracy difference between the best model and the best fair model is 0.015. And the distance is reduced from 0.8320 to 0.2769.

We observe that the distribution of the accuracy of the models reflects the bias distribution in the datasets. Notice that for the unbiased datasets (see experiments 4.1, 4.4, and 4.5) the area in the map for models with higher accuracy (lighter color) is wider compared with the area found in the cases where the dataset is biased (experiments 4.2, 4.3, and 4.6.) For the cases where the data is biased, the most accurate models are concentrated in the area of the map that mirrors the bias in the data (that is, the area closer to the cell that has a positive predicted ratio similar to that present in the data for each protected attribute.)

The method allowed to generate a set of models and compare how the accuracy and the bias distribution changed in connection with the data used to train the model. Additionally, in cases where the most accurate model is located out of the fair zone, we were able to identify a best fair solution that increases fairness to the minimum threshold. Also, we could understand the trade-off in accuracy from selecting the best fair model instead of the most accurate one.
While the accuracy dropped by an average of 2 percentage points between the most accurate model and the best model within the fair zone, the average distance to (1,1) was reduced by 30\%. This, assuring in all cases a model that complies with the \nicefrac{4}{5} rule.

\section{Discussion}

The use of classifier systems and other algorithmic approaches for human resources tasks such as recruitment and promotion is contentious. It could be argued that algorithms should play no part in these decisions, and that they should be solely down to human judgment. At the same time, humans vary widely between them in how they interpret and apply given guidelines, and human judgment can be every bit as biased as algorithmic judgment. One difference is that classifiers (and other algorithms) can be more readily tested because of temporal consistency. We see our work as contributing by helping prospective users to visualize the biases in their classifiers, and better understand the bias landscape of potential classifiers and trade-offs between bias and prediction performance.

A limitation of our current work is the limited availability of public datasets related to bias in the labor market. To evaluate our method more effectively, we used a human capital dataset where bias was manually introduced through sampling, as well as a second dataset commonly used in fairness research. Additional fairness research datasets could be used to further test the method.

The maximum accuracy obtained by the models in our experiment is slightly lower than what could be attained by a sophisticated conventional learning algorithm, but this slight difference could probably be overcome with some further engineering. In this context, it is worth reiterating that conventional machine learning algorithms result in only a single model and no opportunities for bias characterization.

While there are approaches that seek to alleviate biases rather than characterize them, they all have various shortcomings. For example, dominant classes could be undersampled, but that means that not all of the dataset will be used, likely impacting classifier performance and generalization; it also requires that an analysis of bias is done in the first place to understand which instances to undersample.

\section{Conclusion and Future Work}

We showed that CMA-ME can generate multiple models of varying bias, thereby allowing us to visualize the distribution of the accuracy of the models and its relation with the learned bias. In other words, we can understand the bias-performance trade-off. The result of the method is a set of trained models, where the user can choose any of them for deployment. The accuracy of the best-classifying model is comparable, though slightly inferior to, the output of state of the art conventional machine learning methods. The method is fast and would scale to networks of at least tens of thousands of parameters \cite{tjanaka2023training}. 

We had different bias-performance trade-off levels, going from 0.0431 to 0, while the difference in deviation goes from 0.5832 to 0. The correlation between the trade-off and the difference in the deviation between the most accurate model and the model with best accuracy within the fair zone (best fair) is 0.9150. 

The method we have proposed here can be used as-is for training relatively small-size classifiers. These are useful for prediction from a small set of features. The kind of tabular data that is often used in credit scoring systems or hiring support systems are good examples of this. However, bias is common also in ML systems that work on much higher-dimensional, non-tabular data. A good example is face recognition systems, which are ubiquitous but can be highly biased, for example by more readily recognizing people of certain ethnicities~\cite{jain2023zero}. These systems typically build on neural networks with millions of parameters, where evolutionary methods such as CMA-ME largely break down. However, there are QD methods capable of effectively training very large networks, such as the recently proposed Differentiable Quality Diversity (DQD) algorithm~\cite{fontaine2021differentiable}. It would be interesting to apply the methodology proposed here to facial recognition systems.

Fairness is not the only concern when using ML models for important or sensitive decision; another important concern is explainability. Neural networks, while powerful, are unfortunately some of the least explainable ML models. Decision trees, on the other hand, are in principle human-interpretable and can be highly accurate on tabular data. The approach presented in this paper could easily be applied to finding interpretable decision trees by exchanging the CMA-ME algorithm for a version of MAP-Elites suited to tree-structured representations.

\bibliographystyle{splncs04}
\bibliography{mybibliography}

\begin{thebibliography}{10}
\providecommand{\url}[1]{\texttt{#1}}
\providecommand{\urlprefix}{URL }
\providecommand{\doi}[1]{https://doi.org/#1}

\bibitem{albemarle}
Albemarle {P}aper {C}o. v. {M}oody, 422 {U.S.} 405 (1975)

\bibitem{adult_2}
Becker, B., Kohavi, R.: {Adult}. UCI Machine Learning Repository (1996), {DOI}: https://doi.org/10.24432/C5XW20

\bibitem{black2022model}
Black, E., Raghavan, M., Barocas, S.: Model multiplicity: Opportunities, concerns, and solutions. In: Proceedings of the 2022 ACM Conference on Fairness, Accountability, and Transparency. pp. 850--863 (2022)

\bibitem{chen2023comprehensive}
Chen, Z., Zhang, J.M., Sarro, F., Harman, M.: A comprehensive empirical study of bias mitigation methods for machine learning classifiers. ACM Transactions on Software Engineering and Methodology  \textbf{32}(4),  1--30 (2023)

\bibitem{CFR}
{Code of Federal Regulations (CFR)}: Uniform guidelines on employee selection procedures (1978). \url{https://www.ecfr.gov/current/title-29/subtitle-B/chapter-XIV/part-1607?toc=1} (1978), title 29, Subtitle B, Chapter XIV, Part 1607

\bibitem{connecticut}
Connecticut v. {T}eal, 457 {U.S.} 440 (1982)

\bibitem{cully2015robots}
Cully, A., Clune, J., Tarapore, D., Mouret, J.B.: Robots that can adapt like animals. Nature  \textbf{521}(7553),  503--507 (2015)

\bibitem{cully2017quality}
Cully, A., Demiris, Y.: Quality and diversity optimization: A unifying modular framework. IEEE Transactions on Evolutionary Computation  \textbf{22}(2),  245--259 (2017)

\bibitem{fontaine2021differentiable}
Fontaine, M., Nikolaidis, S.: Differentiable quality diversity. Advances in Neural Information Processing Systems  \textbf{34},  10040--10052 (2021)

\bibitem{fontaine2020covariance}
Fontaine, M.C., Togelius, J., Nikolaidis, S., Hoover, A.K.: Covariance matrix adaptation for the rapid illumination of behavior space. In: Proceedings of the 2020 genetic and evolutionary computation conference. pp. 94--102 (2020)

\bibitem{griggs}
Griggs v. {D}uke {P}ower {C}o., 401 {U.S.} 424 (1971)

\bibitem{hajian2016algorithmic}
Hajian, S., Bonchi, F., Castillo, C.: Algorithmic bias: From discrimination discovery to fairness-aware data mining. In: Proceedings of the 22nd ACM SIGKDD international conference on knowledge discovery and data mining. pp. 2125--2126 (2016)

\bibitem{hansen2001completely}
Hansen, N., Ostermeier, A.: Completely derandomized self-adaptation in evolution strategies. Evolutionary computation  \textbf{9}(2),  159--195 (2001)

\bibitem{hellstrom2020bias}
Hellstr{\"o}m, T., Dignum, V., Bensch, S.: Bias in machine learning-what is it good for? In: International Workshop on New Foundations for Human-Centered AI (NeHuAI) co-located with 24th European Conference on Artificial Intelligence (ECAI 2020), Virtual (Santiago de Compostela, Spain), September 4, 2020. pp. 3--10. RWTH Aachen University (2020)

\bibitem{hoffman2019artificial}
Hoffman, S., Podgurski, A.: Artificial intelligence and discrimination in health care. Yale J. Health Pol'y L. \& Ethics  \textbf{19}, ~1 (2019)

\bibitem{huang2022overview}
Huang, C., Zhang, Z., Mao, B., Yao, X.: An overview of artificial intelligence ethics. IEEE Transactions on Artificial Intelligence  \textbf{4}(4),  799--819 (2022)

\bibitem{jain2023zero}
Jain, A., Memon, N., Togelius, J.: Zero-shot racially balanced dataset generation using an existing biased stylegan2. arXiv preprint arXiv:2305.07710  (2023)

\bibitem{li2024triangular}
Li, J., Li, G.: The triangular trade-off between robustness, accuracy and fairness in deep neural networks: A survey. ACM Computing Surveys  (2024)

\bibitem{liem2018psychology}
Liem, C.C., Langer, M., Demetriou, A., Hiemstra, A.M., Sukma~Wicaksana, A., Born, M.P., K{\"o}nig, C.J.: Psychology meets machine learning: Interdisciplinary perspectives on algorithmic job candidate screening. Explainable and interpretable models in computer vision and machine learning pp. 197--253 (2018)

\bibitem{mehrabi2021survey}
Mehrabi, N., Morstatter, F., Saxena, N., Lerman, K., Galstyan, A.: A survey on bias and fairness in machine learning. ACM Computing Surveys (CSUR)  \textbf{54}(6),  1--35 (2021)

\bibitem{meyer2023dataset}
Meyer, A.P., Albarghouthi, A., D'Antoni, L.: The dataset multiplicity problem: How unreliable data impacts predictions. In: Proceedings of the 2023 ACM Conference on Fairness, Accountability, and Transparency. pp. 193--204 (2023)

\bibitem{mouret2015illuminating}
Mouret, J.B., Clune, J.: Illuminating search spaces by mapping elites. arXiv preprint arXiv:1504.04909  (2015)

\bibitem{dataset1}
Möbius: Hr analytics: Employee promotion data. \url{https://www.kaggle.com/datasets/arashnic/hr-ana} (2019)

\bibitem{pessach2022review}
Pessach, D., Shmueli, E.: A review on fairness in machine learning. ACM Computing Surveys (CSUR)  \textbf{55}(3),  1--44 (2022)

\bibitem{pugh2016quality}
Pugh, J.K., Soros, L.B., Stanley, K.O.: Quality diversity: A new frontier for evolutionary computation. Frontiers in Robotics and AI p.~40 (2016)

\bibitem{schwartz2022towards}
Schwartz, R., Schwartz, R., Vassilev, A., Greene, K., Perine, L., Burt, A., Hall, P.: Towards a standard for identifying and managing bias in artificial intelligence, vol.~3. US Department of Commerce, National Institute of Standards and Technology (2022)

\bibitem{speicher2018unified}
Speicher, T., Heidari, H., Grgic-Hlaca, N., Gummadi, K.P., Singla, A., Weller, A., Zafar, M.B.: A unified approach to quantifying algorithmic unfairness: Measuring individual \&group unfairness via inequality indices. In: Proceedings of the 24th ACM SIGKDD international conference on knowledge discovery \& data mining. pp. 2239--2248 (2018)

\bibitem{suresh2021framework}
Suresh, H., Guttag, J.: A framework for understanding sources of harm throughout the machine learning life cycle. In: Proceedings of the 1st ACM Conference on Equity and Access in Algorithms, Mechanisms, and Optimization. pp.~1--9 (2021)

\bibitem{sweeney2013discrimination}
Sweeney, L.: Discrimination in online ad delivery. Communications of the ACM  \textbf{56}(5),  44--54 (2013)

\bibitem{tjanaka2023training}
Tjanaka, B., Fontaine, M.C., Lee, D.H., Kalkar, A., Nikolaidis, S.: Training diverse high-dimensional controllers by scaling covariance matrix adaptation map-annealing. IEEE Robotics and Automation Letters  (2023)

\bibitem{pyribs}
Tjanaka, B., Fontaine, M.C., Lee, D.H., Zhang, Y., Balam, N.R., Dennler, N., Garlanka, S.S., Klapsis, N.D., Nikolaidis, S.: pyribs: A bare-bones python library for quality diversity optimization (2023)

\bibitem{verma2018fairness}
Verma, S., Rubin, J.: Fairness definitions explained. In: Proceedings of the international workshop on software fairness. pp.~1--7 (2018)

\bibitem{wang2011novel}
Wang, P., Weise, T., Chiong, R.: Novel evolutionary algorithms for supervised classification problems: an experimental study. Evolutionary Intelligence  \textbf{4},  3--16 (2011)

\bibitem{watson}
Watson v. {F}ort {W}orth {B}ank i\& {T}rust, 487 {U.S.} 977 (1988)

\bibitem{watson2023multi}
Watson-Daniels, J., Barocas, S., Hofman, J.M., Chouldechova, A.: Multi-target multiplicity: Flexibility and fairness in target specification under resource constraints. In: Proceedings of the 2023 ACM Conference on Fairness, Accountability, and Transparency. pp. 297--311 (2023)

\bibitem{zhang2024fairness}
Zhang, Q., Liu, J., Yao, X.: Fairness-aware multiobjective evolutionary learning. IEEE Transactions on Evolutionary Computation  (2024)

\bibitem{zhang2021fairer}
Zhang, Q., Liu, J., Zhang, Z., Wen, J., Mao, B., Yao, X.: Fairer machine learning through multi-objective evolutionary learning. In: International conference on artificial neural networks. pp. 111--123. Springer (2021)

\bibitem{zhang2022mitigating}
Zhang, Q., Liu, J., Zhang, Z., Wen, J., Mao, B., Yao, X.: Mitigating unfairness via evolutionary multiobjective ensemble learning. IEEE transactions on evolutionary computation  \textbf{27}(4),  848--862 (2022)

\end{thebibliography}

\end{document}